\documentclass{article}
\usepackage{acra}
\usepackage[T1]{fontenc}
\usepackage{graphicx}
\usepackage{amsmath,amssymb}
\usepackage{booktabs}
\usepackage{array}
\usepackage{url}
\usepackage[hidelinks]{hyperref}
\usepackage{orcidlink}

\title{Minimum Sufficient Communication for Independent Behaviour Tree Robot Teams}
\author{
    Sherry Jiang\, \orcidlink{0009-0002-0322-4010}
    \and Oliver Obst\,\orcidlink{0000-0002-8284-2062} \\
    School of Computer Science and Engineering \\
    UNSW Sydney, Australia \\
    \texttt{sherry.jiang@student.unsw.edu.au}\quad
    \texttt{o.obst@unsw.edu.au}
}

\begin{document}
\maketitle

\begin{abstract}
    We present an experimental framework for determining sufficient communication in
    robot teams running fixed, independent Behaviour Trees with private blackboards.
    Within a declared policy family, we vary message content and admission frequency
    and measure traffic in recipient-weighted bits per robot decision. A dense sweep
    identifies candidate policies for comparison with full-state communication. We
    retest selected candidates and the reference on fresh, paired scenarios, using
    one-sided simultaneous confidence bounds to assess whether success remains
    within five percentage points of the full-state reference. The bounds also
    assess whether tested cheaper policies can be ruled out as insufficient.
    Depending on the evidence, the framework reports an exact minimum, a positive
    bracket, a sufficient upper bound, or an unresolved boundary. Experiments with
    eight content policies across four warehouse conditions identify zero
    communication as the exact minimum under low contention. In the scaled
    condition, the no-release policy at gate $q=0.992222$ is confirmed sufficient,
    achieving 96.33\% success at an estimated rate of 738.55 bits per robot
    decision, or 0.544\% of full-state traffic. The positive minimum remains
    unresolved. These claims apply only to the declared policy family, sampled gate
    values and operating conditions.
\end{abstract}

\section{Introduction}

Communication allows a robot to learn that another robot has already claimed or
completed a task, which can prevent repeated travel and work. However, sending
this information is associated with a cost and creates communication traffic.
Therefore, the challenge is to reduce communication while preserving enough
information for the robot team to coordinate successfully.

We therefore investigated how much communication is sufficient, for a fixed team
controller and operating condition, to keep task success within a chosen loss
from full-state communication. We study this question using Behaviour Trees
(BTs), which provide modular and reactive control structures
\cite{colledanchise2018,iovino2022}. Each robot runs its own BT instance and
maintains private beliefs about tasks and teammates. Throughout the experiment,
the BT structure and task-selection rule remain unchanged. We vary only the
communication protocol, which determines what information is shared between
robots and how ownership information is maintained. This allows us to isolate
the effect of communication while keeping the underlying decision-making process
fixed.

We vary communication along two dimensions, namely message content and admission
probability. Firstly, a content rule determines what information a policy is
allowed to send and when it is generated, such as task completion, ownership
claims, releases, intent, task facts or robot status. Secondly, a gate
\(q\in[0,1]\) controls the probability that each eligible message is admitted.
The gate therefore controls how often eligible messages are transmitted, but it
does not directly measure the amount of communication. Specifically, different
content rules generate different numbers and sizes of messages, and messages
sent earlier can change later robot decisions and therefore which messages are
generated afterwards. We therefore measure the traffic that is actually
transmitted. For each message, we count its application-level payload once for
every intended recipient, then normalise the total by the number of robot
decisions. This gives the realised recipient-weighted application bits per robot
decision.

A policy is sufficient if its success remains within the chosen tolerance of the
full-state reference. To claim a positive minimum, we would additionally need
evidence that every cheaper policy in the declared family is insufficient. We
therefore distinguish between a confirmed sufficient policy, an unresolved
boundary, a positive bracket, and an exact zero-rate minimum.

Note that our conclusions apply only to the family that we evaluate. Within this
scope, our contribution is a reproducible procedure that combines fixed private
BT control, measured recipient-weighted cost, exploratory frontier search, and
fresh paired confirmation. We apply the procedure to eight content policies
across four controlled task conditions and report the strongest conclusion
supported by the evidence in each case.

\section{Related Work and Gap}

Early multi-robot research showed that communication requirements depend
strongly on the task. Communication can improve performance in some settings
while being unnecessary in others, and relatively simple communication can
sometimes approach richer schemes \cite{balch1994}. Task-allocation research
similarly shows that the usefulness of information depends on the structure of
the team and the way tasks are coupled \cite{gerkey2004,dias2006}. This
motivates studying a framework to test on specific controller and operating
condition, rather than searching for one universal minimum across unrelated
robot teams.

Behaviour Tree research has developed several ways to coordinate multiple
robots. Distributed BT execution, capability-based allocation and communication
of robot intent have all been used to improve coordination
\cite{heppner2023,heppner2024,hull2024}. These approaches mainly change the
coordination architecture or the information representation used by the team. In
contrast, we keep the local BT structure and task-selection rule fixed and vary
only the communication protocol. This allows us to study how changing the
information shared between robots affects the performance of the same underlying
controller.

Furthermore, previous work has used utility-aware message forwarding, learned
communication schedules and information bottlenecks to decide what information
should be transmitted \cite{wang2020}. Our contribution is therefore not simply
to reduce communication, but to provide an experimental and statistical
procedure for determining when communication is sufficient. The resulting
frontier relates communication cost to task performance, similar in spirit to
rate-distortion analysis. However, it describes only the tested policy family
and is not an information-theoretic lower bound.

\section{Proposed Experimental Framework}

\subsection{Setup and definitions}

An operating condition $\theta$ defines the fixed experimental setting. It
specifies the number of robots and tasks, the task distribution, arena size,
time horizon, capability assignment, sensing assumptions, and channel delay and
loss. Within a condition, a scenario seed determines the task locations and the
reproducible pseudo-random values used by the communication gate. Robot starting
positions remain fixed. We treat one complete team episode as one experimental
observation.

A communication policy is written as $c=(S,q)$. The content rule $S$ specifies
what information can be sent, when each type of message is generated, and how
long ownership information remains valid. The gate $q\in[0,1]$ controls whether
an eligible message is admitted for transmission. Every admitted message is
addressed to all other robots, while the channel conditions are fixed by
$\theta$.

For a scenario $\omega$, let $Y(c,\omega)=1$ if all tasks are completed by the
end of the $H$-tick episode, and $Y(c,\omega)=0$ otherwise. The success
probability of policy $c$ under condition $\theta$ is therefore given by

\begin{equation}
    p_\theta(c)
    =
    \Pr\{Y(c,\omega)=1\mid\theta\}.
    \label{eq:success}
\end{equation}

Success is the primary performance measure. Completion fraction and duplicate
work are reported as secondary measures to help explain changes in success.
Makespan is reported only for successful episodes.

Communication cost is measured from the messages that are actually admitted for
transmission. For a message $m$, let $L(m)$ be the length of its serialised
payload in bytes and let $\mathcal R(m)$ be its intended recipients. The total
application-level communication cost in scenario $\omega$ is

\begin{equation}
    B_{\mathrm{app}}(c,\omega)
    =
    \sum_{m\text{ admitted}}
    8L(m)|\mathcal R(m)|.
    \label{eq:bits}
\end{equation}

All episodes run for the full horizon, even if every task finishes early. With
$N$ robots and $H$ ticks, each episode therefore contains a fixed $NH$ robot
decision opportunities. We define the communication rate as

\begin{equation}
    R_\theta(c)
    =
    \frac{
    \mathbb E[B_{\mathrm{app}}(c,\omega)\mid\theta]
    }{NH}.
    \label{eq:rate}
\end{equation}

The rate is measured in application bits per robot decision. A message is
charged when it is transmitted, so later network loss does not remove its
communication cost. Packets scheduled for delivery after the episode horizon
also count towards transmitted traffic, although they do not affect the episode
outcome.

\subsection{Sufficiency and minimum}

We compare each policy with a full-state reference, denoted $c_{\mathrm{ref}}$.
Full-state communication uses the same Behaviour Trees and local sensing as
every other policy, but broadcasts all private task records and robot status
information at every tick. Using this reference allows us to compare
communication policies while keeping the underlying decision-making process
unchanged.

Before confirmation, we choose a non-inferiority margin of $\delta=0.05$. A
policy is considered sufficient if its success probability is no more than five
percentage points below that of the full-state reference, i.e., a policy $c$ is
sufficient when

\begin{equation}
    \Delta_\theta(c)
    =p_\theta(c)-p_\theta(c_{\mathrm{ref}})
    \geq-\delta.
    \label{eq:sufficiency}
\end{equation}

The value $\delta=0.05$ is a design choice, representing the performance loss we
are willing to accept for this experiment. More generally, this could be altered
depending on the specific scenarios.

Therefore, for a declared policy family $\mathcal C$, the communication rate we
would identify is given by

\begin{equation}
    R^*_{\theta,\delta,\mathcal C}
    =
    \inf
    \left\{
    R_\theta(c):
    c\in\mathcal C,\;
    \Delta_\theta(c)\geq-\delta
    \right\}.
    \label{eq:minimum}
\end{equation}

To compare performance across different communication budgets, we define the
best-achievable budget frontier

\begin{equation}
    F_\theta(r)
    =
    \sup_{
    c\in\mathcal C:
    R_\theta(c)\leq r
    }
    p_\theta(c).
    \label{eq:frontier}
\end{equation}

For a communication budget $r$, the frontier gives the highest success
probability achievable by any policy whose communication rate is at most $r$. In
the experiment, we approximate this by ordering the tested policies from lowest
to highest measured rate and, at each rate, recording the highest success
observed so far. In particular, if a more expensive policy performs worse than a
cheaper one, it does not lower the frontier because the cheaper policy is still
available within that budget. The empirical frontier therefore stays constant or
increases as the communication budget grows.

We then use this frontier to identify low-rate policies that appear promising
and retest them on a different set of scenarios. If a retested policy $c$ is
confirmed to be sufficient, then we know that sufficient performance can be
achieved using communication rate $R_\theta(c)$. Therefore,

\begin{equation}
    R^*_{\theta,\delta,\mathcal C}
    \leq
    R_\theta(c).
\end{equation}

This gives an upper bound on the minimum sufficient rate - the true minimum
could be lower, but it cannot be higher than the rate of a policy already shown
to be sufficient.

The observed mean rate $\widehat R$ is an estimate of $R_\theta(c)$ rather than
an exact bound on the population minimum. Because it is calculated from a finite
number of episodes, it has sampling uncertainty, which we report separately.

\subsection{Interpreting the evidence}

We use the frontier to identify policies that achieve strong performance at low
communication rates. These selected policies are then tested again on different
scenarios that were not used during the exploratory stage. Let
$\Delta(c)=p_{\theta}(c)-p_{\theta}(c_{\mathrm{ref}})$. A candidate policy at
rate $R_U$ is confirmed sufficient when its simultaneous one-sided lower
confidence bound satisfies $ \mathrm{LCB}\{\Delta(c_U)\}\geq-\delta.$ Passing
this test shows that sufficient performance can be achieved at rate $R_U$.
Therefore, $R^* \leq R_U.$ However, this does not show that $R_U$ is the
minimum, because a cheaper sufficient policy may still exist.

A stronger conclusion is possible only if all cheaper policies can also be ruled
out. If every tested policy with rate at or below $R_L$ is shown to be
insufficient, while a policy at $R_U$ is confirmed sufficient, then the minimum
is constrained to $R_L < R^* \leq R_U.$ This gives a positive bracket on the
minimum at the resolution of the tested rates.

Zero communication is a special case. If the zero-rate policy is confirmed
sufficient, then $R^*=0.$

\section{Experiment and Implementation}

\subsection{Task Environment and Robot Behaviour}

The experiment uses a two-dimensional grid containing stationary service tasks.
There are no pickup or delivery stages, obstacles, collisions, deadlines, or
battery constraints. All tasks are available from the start of an episode. Each
robot initially knows every task's identifier, location, and required
capability. This initial task catalogue is treated as part of the problem setup
and is not counted as communication. The experiment focuses on communication
used for coordination after the tasks are known.

Task locations are sampled independently and uniformly from cells in the grid,
and multiple tasks may occupy the same cell. Capabilities are assigned cyclically
to robots and tasks, and a robot may only work on compatible tasks. Robot starting
positions are fixed in a central $2\times2$ block; the eight-robot condition
reuses these four cells. Perfect local sensing within a Manhattan radius of one
grid cell reveals task completion. Teammate status is available only through
messages. Table~\ref{tab:conditions} gives the four operating conditions.

\begin{table}[t]
    \centering
    \small
    \setlength{\tabcolsep}{3pt}
    \begin{tabular}{@{}lrrrrrr@{}}
        \toprule
        Condition & $N$ & Tasks & $H$ & $A$ & $K$ & Delay/loss \\
        \midrule
        Baseline & 4 & 12 & 50 & 16 & 2 & 1 / 0\% \\
        Low contention & 4 & 12 & 50 & 16 & 4 & 1 / 0\% \\
        Lossy & 4 & 12 & 50 & 16 & 2 & 2 / 10\% \\
        Scaled & 8 & 24 & 65 & 22 & 2 & 1 / 0\% \\
        \bottomrule
    \end{tabular}
    \caption{Operating conditions: $N$ robots, horizon $H$ ticks, an $A\times A$
    grid and $K$ capability types. The last column gives transmission-to-delivery
    delay in ticks and loss probability per recipient.}
    \label{tab:conditions}
\end{table}

A robot selects the nearest compatible task that it believes is unfinished and
not already held by another active claim. Distance ties are broken using the
task identifier. If teammate status information is available, the robot may also
defer to a compatible teammate that it believes is closer to the task, has
priority under the tie-breaking rule, or is already working on that task.

Task selection takes one tick. On later ticks, the robot moves one grid cell
towards its selected task, moving along the $x$ direction before the $y$
direction. Once it reaches the task, each work tick contributes one unit of
progress. Every task requires two units of work. If several robots work on the
same task during the same tick, only one unit of progress is added, and the
remaining actions are counted as duplicate work.

The low-contention condition assigns one capability type to each of the four
robots. Each task therefore has only one eligible robot, removing competition
between robots with the same capability. The scaled condition changes the team
size, number of tasks, arena size, and horizon together. It would be interpreted
as a different operating condition.

\subsection{Private Behaviour Trees and Tick Order}

Each robot has its own Behaviour Tree instance, blackboard, task beliefs, and
teammate beliefs. All robots use the same fixed BT structure with three
branches: continue an active task; otherwise select a new task; otherwise remain
idle. The environment stores the true task state and applies the robots'
actions, but it does not choose tasks for them.

Each simulation tick follows the same order. First, messages scheduled for that
tick are delivered and local sensing is updated. Each robot then ticks its own
BT using only its private information. Robot actions are committed, and outgoing
communication is generated afterwards.

A packet transmitted at tick $t$ is delivered at tick $t+d$, where $d$ is the
communication delay for that condition. Messages generated during a tick
therefore cannot affect another robot until a later tick. This ensures that
communication timing is determined by the model, rather than by the order in
which robots happen to be processed by the simulator.

\subsection{Communication Policies}

The experiment evaluates eight communication policies. Table~\ref{tab:policies}
outlines each of the eight policies. Each policy determines which types of
information may be sent. The underlying BT and task-selection rule remain
unchanged.

A completion message is generated when a task finishes. A claim is generated
when a robot selects a task, and a release is generated when a robot gives up a
task because of a competing claim. Some policies use persistent claims, while
others use leases that expire unless renewed. Lease duration is eight ticks,
or ten in the scaled condition. An active owner renews its private lease each
tick; peers learn of renewal only through delivered messages.

\begin{table*}[t]
    \centering
    \small
    \begin{tabular}{@{}
        >{\raggedright\arraybackslash}p{\dimexpr0.20\textwidth-1.5\tabcolsep\relax}
        >{\raggedright\arraybackslash}p{\dimexpr0.29\textwidth-1.5\tabcolsep\relax}
        >{\raggedright\arraybackslash}p{\dimexpr0.20\textwidth-1.5\tabcolsep\relax}
        >{\raggedright\arraybackslash}p{\dimexpr0.31\textwidth-1.5\tabcolsep\relax}
        @{}}
        \toprule
        Policy & Eligible messages & Ownership & Periodic content\\
        \midrule
        Completion
        & Completion
        & No exchanged claims
        & None\\
        Claim--release
        & Completion, claim, release
        & Persistent
        & None\\
        Lease--heartbeat
        & Above + intent
        & Leased
        & Intent every active tick\\
        Rich status
        & Above + task fact, status
        & Leased
        & One task fact and status each tick\\
        No claim
        & Completion, release
        & No exchanged claims
        & None\\
        No release
        & Completion, claim
        & Persistent
        & None\\
        Lease, no heartbeat
        & Completion, claim, release
        & Leased
        & None\\
        Rich, no status
        & All rich content except status
        & Leased
        & Intent and one task fact each tick\\
        \midrule
        Full-state reference
        & All message types
        & Leased
        & Intent, all task facts and status each tick\\
        \bottomrule
    \end{tabular}
    \caption{Communication policies used in the experiment. All proposed messages
    are addressed to the other $N-1$ robots.}
    \label{tab:policies}
\end{table*}

Note that messages are serialised using a fixed text format. Their exact byte
length is measured after serialisation rather than assigning one fixed size to
each message type. This matters because optional fields and different identifier
lengths can change the size of individual messages.

\subsection{Admission Gate and Message Loss}

For every message that is eligible under a policy, the simulator generates a
reproducible pseudo-random score $u_m\in[0,1).$ The message is admitted when
$u_m<q$, where $q\in[0,1]$ is the admission gate. Thus $q=0$ admits no eligible
messages, $q=1$ admits all of them, and intermediate values admit only a
fraction of the available message opportunities.

The score for a message depends on the scenario seed and the semantic details of
the message, including its type, sender, tick, task, owner, and original claim
time. This makes admission reproducible across repeated runs. If the same
message opportunity exists at two different gate values, increasing $q$ cannot
cause that particular message to change from admitted to rejected.

However, increasing $q$ can also change the robots' later behaviour. For
example, admitting an ownership claim may cause another robot to choose a
different task, which can remove message opportunities that would otherwise have
occurred later in the episode. Therefore, a larger $q$ does not simply mean that
the final set of transmitted messages is a larger version of the set produced at
a smaller $q$. For this reason, $q$ is used only to control message admission.
It is not used as the communication cost itself. Instead, we calculate the
realised communication rate from the messages that are actually transmitted
during the episode.

Messages rejected by the gate are discarded. Messages admitted by the gate are
transmitted and are then subject to the channel-loss model separately for each
recipient. The main simulation does not include retransmissions,
acknowledgements, congestion, or communication-range limits.

\subsection{Staged Experimental Design}

We design the experiments such that early stages search a large number of
policy--gate combinations to find promising low-communication candidates. These
candidates are then tested again using new scenario seeds that were not used
during selection.

\begin{table*}[t]
    \centering
    \small
    \begin{tabular}{@{}
        >{\raggedright\arraybackslash}p{\dimexpr0.14\textwidth-1.6\tabcolsep\relax}
        >{\raggedright\arraybackslash}p{\dimexpr0.39\textwidth-1.6\tabcolsep\relax}
        >{\raggedleft\arraybackslash}p{\dimexpr0.10\textwidth-1.6\tabcolsep\relax}
        >{\raggedleft\arraybackslash}p{\dimexpr0.14\textwidth-1.6\tabcolsep\relax}
        >{\raggedright\arraybackslash}p{\dimexpr0.23\textwidth-1.6\tabcolsep\relax}
        @{}}
        \toprule
        Stage & Design & Episodes & First seed & Role\\
        \midrule
        Screen
        & 8 policies $\times$ 21 gates; two anchors
        & 300
        & 6,000,001
        & Exploration\\
        Detailed
        & Retained policies $\times$ 101 gates; anchors
        & 400
        & 7,000,001
        & Exploration\\
        Refinement
        & Local gates and endpoints
        & 1,000
        & 8,000,001
        & Candidate selection\\
        Confirmation
        & Selected neighbours, guards and anchors
        & 3,000
        & 9,000,001
        & Primary inference\\
        Token test
        & 3 refill factors $\times$ 2 capacities
        & 1,000
        & 10,000,001
        & Secondary stress test\\
        \bottomrule
    \end{tabular}
    \caption{Experimental stages. Different seed blocks are used at each stage.
    Within a stage and condition, all policies are evaluated on the same scenario
    seeds. The corrected experiment contains 2,119,600 primary episodes and
    24,000 token-test episodes.}
    \label{tab:stages}
\end{table*}

The screen evaluates all eight content policies over 21 gate values. Policies
that remain close to the reference performance target are retained for the
detailed stage, where 101 equally spaced gate values are tested. At least two
content policies are retained in each condition.

The refinement stage adds 30 gate values around up to three promising regions
for each retained policy, together with the endpoints. These exploratory stages
are used only to locate candidate operating points.

Before confirmation results are examined, the policies to be retested are fixed.
The confirmation set includes the proposed candidate, a cheaper challenger, a
nearby higher-rate policy, and additional higher gate values for the proposed
content rule. The zero-message and full-state policies are also included. Once
this confirmation plan is fixed, no new policy is added based on the
confirmation results.

All stages use separate seed blocks. Within a stage, however, every policy and
the full-state reference use the same scenario seeds. This allows their outcomes
to be compared on exactly the same task layouts and channel realisation
structure.

\subsection{Paired Confirmation and Confidence Bounds}

The final sufficiency decision uses the fresh confirmation scenarios. For policy
$c$ and scenario $i$, define

\begin{equation}
    d_i(c)
    =
    Y_i(c)-Y_i(c_{\mathrm{ref}}).
    \label{eq:paired}
\end{equation}

The paired difference can take three values. A value of $1$ means the candidate
succeeds while the reference fails, $-1$ means the candidate fails while the
reference succeeds, and $0$ means that both have the same outcome.

Pairing makes the comparison fair by evaluating both policies on the same
scenarios. This reduces the effect of some scenarios being harder than others.
If both policies succeed or both fail on the same scenario, that scenario
contributes zero to their performance difference because neither policy
performed better than the other.

Let
\[
a(c)=\Pr\{d_i(c)=1\}
\]
be the probability that the candidate gains over the reference, and let
\[
b(c)=\Pr\{d_i(c)=-1\}
\]
be the probability that it loses to the reference. The success difference is

\begin{equation}
    \Delta(c)=a(c)-b(c).
\end{equation}

We construct one-sided Clopper-Pearson bounds for these two probabilities. Let
$M$ be the total number of non-reference policies included in confirmation
across all conditions, and define

\begin{equation}
    \eta=\frac{0.05}{2M}.
\end{equation}

If $n_+$ is the number of paired gains and $n_-$ is the number of paired losses,
the simultaneous lower bound for the success difference is

\begin{equation}
    L(c)
    =
    L_{\mathrm{CP}}(n_+,n;\eta)
    -
    U_{\mathrm{CP}}(n_-,n;\eta).
    \label{eq:exact}
\end{equation}

For $0<k<n$,

\begin{align}
    L_{\mathrm{CP}}(k,n;\eta)
    &=
    Q_{\eta}
    \{\mathrm{Beta}(k,n-k+1)\},\\
    U_{\mathrm{CP}}(k,n;\eta)
    &=
    Q_{1-\eta}
    \{\mathrm{Beta}(k+1,n-k)\}.
\end{align}

At $k=0$, the bounds are $L_{\mathrm{CP}}=0$ and
$U_{\mathrm{CP}}=1-\eta^{1/n}$; at $k=n$, they are
$L_{\mathrm{CP}}=\eta^{1/n}$ and $U_{\mathrm{CP}}=1$.
With no paired gains or losses, Equation~\ref{eq:exact} therefore gives
$L(c)=\eta^{1/n}-1$, approximately $-0.0023$ for $M=25$ and $n=3000$.

The adjustment by $M$ accounts for the fact that several confirmation candidates
are tested at the same time. Using a union bound gives at least 95\%
simultaneous coverage for the family of lower bounds, assuming the scenario
samples themselves are independent.

A policy is confirmed sufficient when $L(c)\geq-0.05.$ This means that, after
accounting for uncertainty and multiple comparisons, the evidence supports the
policy being no more than five percentage points worse than the full-state
reference.

We inspect every policy included in confirmation and report the passing policy
with the lowest measured mean communication rate.

We also report 95\% confidence intervals for absolute success rates and mean
communication rates to show the uncertainty in these estimates.

\subsection{Exploratory Frontier Uncertainty}

As discussed previously, the detailed-stage frontier is exploratory and is used
to help select candidate policies. To show its uncertainty, we use a paired
bootstrap with 2,000 replicates \cite{davison1997}. Each bootstrap replicate
resamples complete scenarios, using the same resampled scenario indices for all
policies within a condition.

For each replicate, we calculate the largest estimation error across the tested
policies. These errors are used to construct a common one-sided uncertainty
envelope around the exploratory frontier. The resulting shaded region in
Figure~\ref{fig:frontiers} is used to visualise uncertainty during candidate
selection.

\subsection{Secondary Token-Bucket Test}

As a secondary stress test, we also replace probability gating with a
token-bucket communication limit. For this test, $q=1$, so all eligible messages
are proposed, but a message can only be transmitted when the robot has enough
tokens available.

The bucket is tested using refill rates equal to $0.9$, $1.0$, and $1.1$ times
the proposed candidate's estimated communication rate. Each robot's bucket
starts empty. Capacities are 16,384 or 65,536 bits, doubled in the scaled
condition. Messages that cannot be sent immediately may remain pending, while
periodic messages are regenerated.

This test is not part of the primary sufficiency analysis. Its purpose is to
check whether policies with similar average communication rates behave similarly
when the timing and burstiness of communication are constrained.

\section{Results}

\begin{figure*}[p]
    \centering

    \includegraphics[
        width=\textwidth,
        height=.41\textheight,
        keepaspectratio
    ]{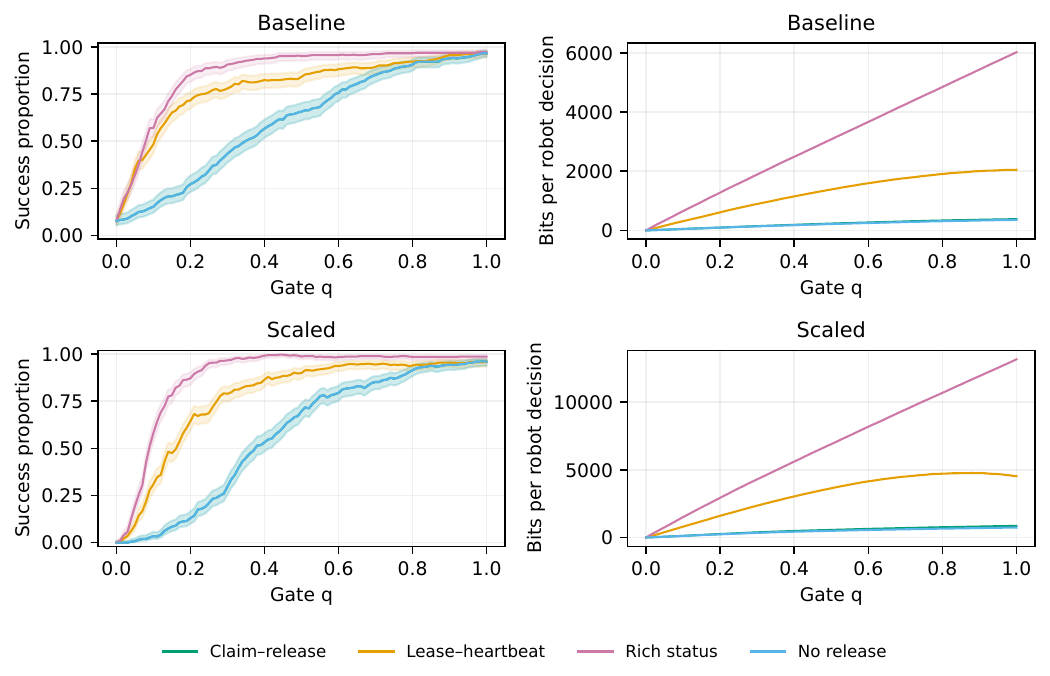}
    \caption{Gate response in baseline and scaled operation for four selected content rules. Left: observed success with pointwise 95\% Wilson intervals. Right: measured mean application rate.}
    \label{fig:qrate}

    \includegraphics[
        width=\textwidth,
        height=.5\textheight,
        keepaspectratio
    ]{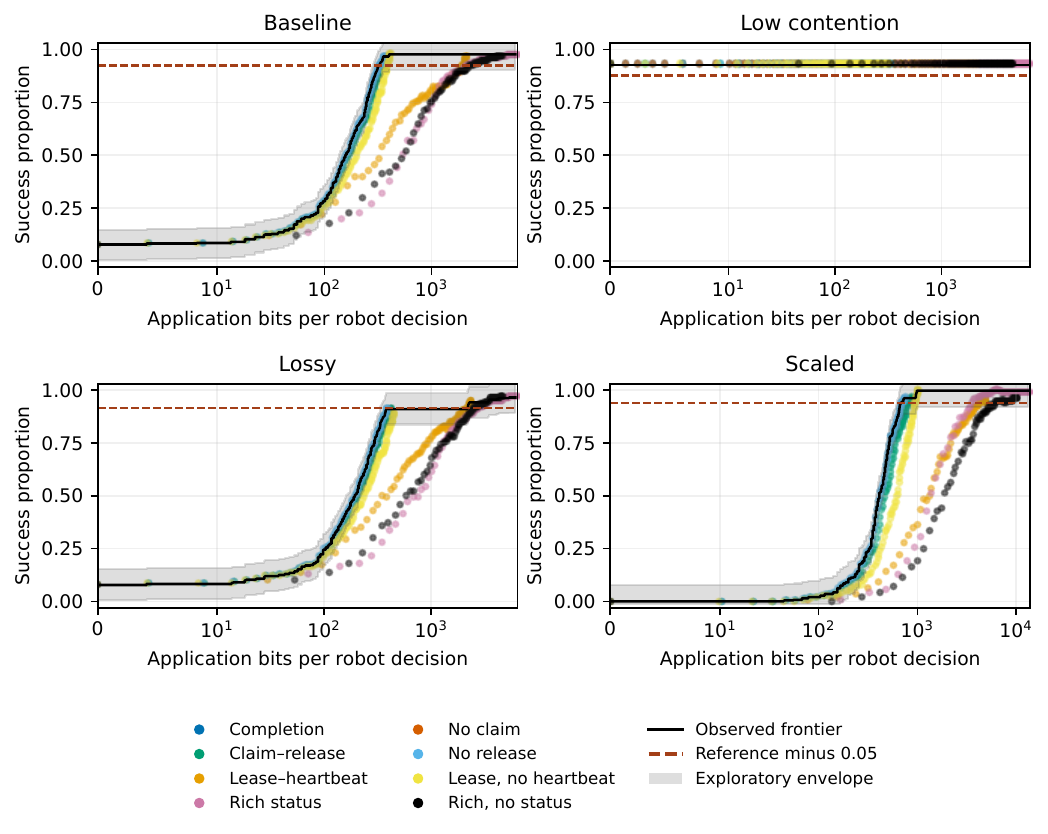}
    \caption{Detailed-stage frontiers with an explicit policy-colour legend. Each point is a tested arm and the black step is the running maximum of observed success. The dashed line is the observed full-state success minus 0.05. Shading is the running envelope of paired-bootstrap bounds shifted by the observed reference mean. The horizontal axis uses a symmetric logarithmic scale with a linear region near zero.}
    \label{fig:frontiers}

\end{figure*}

\begin{table*}[t]
    \centering
    \small
    \begin{tabular}{llrrrrr}
\toprule
Condition & Policy, $q$ & Ref. success \% & Success \% & $\widehat R$ (bits/decision) & $L$ & $\rho$ \\
\midrule
Baseline & \shortstack[l]{No release\\$q=1$} & \shortstack{97.33\\(96.69, 97.85)} & \shortstack{95.40\\(94.59, 96.09)} & \shortstack{356.36\\(355.98, 356.74)} & $-0.0395$ & 1.133\% \\[4pt]
Low contention & \shortstack[l]{No messages\\$q=0$} & \shortstack{93.37\\(92.42, 94.20)} & \shortstack{93.37\\(92.42, 94.20)} & \shortstack{0.00\\(0.00, 0.00)} & $-0.0023$ & 0.000\% \\[4pt]
Lossy & \shortstack[l]{Rich status\\$q=0.4989$} & \shortstack{96.60\\(95.89, 97.19)} & \shortstack{93.67\\(92.74, 94.48)} & \shortstack{3127.66\\(3121.17, 3134.15)} & $-0.0461$ & 9.902\% \\[4pt]
Scaled & \shortstack[l]{No release\\$q=0.9922$} & \shortstack{99.10\\(98.69, 99.38)} & \shortstack{96.33\\(95.60, 96.95)} & \shortstack{738.55\\(737.50, 739.60)} & $-0.0438$ & 0.544\% \\[4pt]
\bottomrule
\end{tabular}

    \caption{Cheapest passing arm among all retested candidates in each condition. Parentheses give pointwise 95\% Wilson intervals for success and descriptive 95\% intervals for the mean rate. $L$ is the simultaneous paired lower bound from Equation~\ref{eq:exact}; passing requires $L\geq-0.05$. The ratio $\rho=\widehat R(c)/\widehat R(c_{\mathrm{ref}})$ compares the mean confirmation communication rates of the candidate and full-state reference within the same condition, expressed as a percentage.}
    \label{tab:confirmation}
\end{table*}

\subsection{Content and gate response}

Figure~\ref{fig:qrate} shows that increasing the gate generally increases the
realised communication rate, but higher traffic alone does not guarantee
higher success. Performance depends strongly on which information is being
communicated.

In the baseline condition, completion-only communication achieved 13.00\%
success, whereas claim--release achieved 97.67\%. Mean duplicate work also fell from 14.67 to 0.00 actions per episode. This suggests that learning only when a task has finished is often too late to prevent redundant behaviour. Ownership claims are more useful because they can prevent another robot from travelling towards the same task before the work is completed. Removing release messages still gave 97.67\% success at 354.55 bits per robot decision. These comparisons suggest that claims provide most of the useful coordination information in this condition.

The same pattern becomes stronger in the scaled condition. Completion-only communication achieved just 1.00\% success, while claim-release achieved 95.67\%. Mean duplicate work fell from 75.95 to 0.00 actions per episode. Removing release messages again retained 95.67\% success, at 742.68 bits per robot decision. This indicates that, as more robots can compete for the same tasks, communicating ownership becomes increasingly important for avoiding duplicate work. At the same time, the strong performance of the no-release policy suggests that not every ownership-related message is equally important.

The lossy condition shows a different communication requirement. Persistent claim-release achieved 92.00\% success, lease-heartbeat achieved 96.33\%, and leased claims without heartbeat achieved 89.33\%. The improvement from adding heartbeat messages suggests that repeated ownership information can help when earlier messages may be lost. This suggests that useful communication policy depends on the operating condition: a sparse event-based policy can work well in a reliable channel, while message loss can make repeated updates more valuable.

Figure~\ref{fig:frontiers} combines all detailed-stage policy--gate
combinations on a common rate-success plot. The frontier records the best
observed success available up to each communication rate. It therefore shows the trade-off between communication cost and performance across policies.

\subsection{Fresh-seed confirmation}

The exploratory stages identify promising operating points, but final sufficiency claims use new confirmation scenarios that were not used to select those candidates. Table~\ref{tab:confirmation} reports the lowest-rate passing arm among the policies retested in each condition. The simultaneous family contains 25 non-reference arms across all four conditions, with 3000 paired scenarios per arm. 

In the baseline condition, the no-release policy at $q=1.0000$ achieved
95.40\% success at an estimated 356.36 bits per robot decision, or 1.133\% of full-state traffic. Its simultaneous paired lower bound was $-0.0395$, so it passes the $-0.05$ sufficiency criterion. This confirms that, for the baseline condition, completion and claim messages are sufficient without release messages. 

The low-contention condition gives a stronger result. The zero-message policy achieved 93.37\% success, matching the full-state reference on the paired sample, with a lower bound of $-0.0023$. Its communication rate is therefore 0.00 bits per robot decision. Together with the structural argument in Section~\ref{sec:zero}, this establishes that communication is unnecessary
under this condition: because every task capability has only one eligible
robot, the robots do not need ownership messages to avoid competing for the
same task.

The lossy condition required substantially richer communication among the
retested candidates. Rich status at $q=0.4989$ achieved 93.67\% success at
3127.66 bits per robot decision, or 9.902\% of full-state traffic, with a
lower bound of $-0.0461$. It therefore passes the sufficiency criterion.
Combined with the exploratory heartbeat results, this suggests that when
messages can be lost, repeated state information can compensate for missing earlier coordination messages. Note that this result identifies a sufficient operating point, not the minimum communication required under loss.

In the scaled condition, the no-release policy at $q=0.9922$ achieved
96.33\% success at 738.55 bits per robot decision. This is only 0.544\% of the full-state communication rate, and its simultaneous lower bound of
$-0.0438$ passes the sufficiency criterion. The result shows that the scaled team can remain within the chosen success tolerance using a sparse
claim-based protocol rather than repeatedly broadcasting full task and robot state.

The scaled condition also illustrates why fresh confirmation is necessary. The exploratory procedure proposed the no-release policy at $q=0.9422$. When retested on fresh scenarios, it achieved 94.70\% success at 717.58 bits per robot decision, but its lower bound was $-0.0621$, so it did not pass the confirmation criterion. The selected cheaper challenger at $q=0.9300$ similarly achieved 94.70\% success at 712.25 bits per robot decision and did not pass. The confirmed operating point therefore moved to the higher gate $q=0.9922$. This shows that an apparent crossing in the exploratory frontier is not sufficient.

\subsection{Token-bucket stress test}

The scaled token-bucket tests produced mean communication rates between
571.99 and 667.41 bits per robot decision, but success remained between
2.30\% and 11.40\%. These rates are of the same order as the probability-gated
candidate, yet their performance is much worse.

This result shows that average communication rate alone does not determine
performance. The token bucket changes when messages can be sent and may delay
or suppress ownership information that is useful at a particular decision
point. Two mechanisms can therefore produce similar average traffic while
delivering different information at different times.

\subsection{Why zero communication is sufficient}
\label{sec:zero}

In the low-contention condition, each task has exactly one eligible robot.
Messages about other robots' tasks cannot remove a compatible option, and
status-based deferral considers only compatible teammates, of which there are
none. Tasks do not interact through collisions or shared resources. The
compatible-task trajectory is therefore identical with and without
communication, by induction over ticks. Hence $p(c_0)=p(c_{\mathrm{ref}})$ for
every task layout under these assumptions, where $c_0$ is the zero-message
policy, and $R^*=0$. The paired experiment checks the implementation; identical
outcomes in a finite sample alone would not establish population equality.

\section{Discussion}

\subsection{Answer to the research question}

There is no single communication minimum for all independent BT teams. The
quantity is a function of the task condition $\theta$, acceptable loss $\delta$,
and policy family $\mathcal{C}$. Within those boundaries, the framework can make
rigorous statements by holding the controller fixed, searching content and
frequency separately, measuring actual bits, and applying paired uncertainty
bounds.

We identify an exact zero-rate minimum under low contention. In baseline and
scaled operation, the no-release policy is confirmed sufficient at $q=1$ and
$q=0.992222$, respectively. Under loss, rich status at $q=0.4989$ passes.
Confirmation does not exclude every cheaper policy, so the positive minima
remain unresolved.

For BT engineering, the frontier turns an informal design process into a
measurable test. A developer can ask whether adding a message type moves the
frontier left or merely adds cost. The ablations also yield concrete domain
findings. Claims are the smallest tested semantic component that removes most
duplicate work. Releases are redundant under deterministic task completion.
Under loss, renewable ownership becomes useful. Because the BT is unchanged,
these effects can be traced to information entering the tree rather than to a
hidden change in decision logic.

\subsection{Novelty and limits}

Our contribution is a method for assessing communication requirements with a
fixed controller. Prior work designs more capable BT coordination or learned
sparse protocols. We ask which content justifies its communication cost, then
assess the smallest measured rate the data support. Private BT blackboards
provide a controlled information boundary, within which we assess the
content--frequency frontier using the explicit claim hierarchy.

Four limits remain. First, the minimum is only over eight declared content
policies and sampled gate values. A better encoding may use fewer bits. Second,
all conditions are warehouse variants. Adaptability must be tested in a domain
with a different reason for communication, such as partial-discovery search and
rescue or tasks requiring simultaneous coalitions. Third, full state is an
information reference using the same BT, not a centrally optimal allocator.
Fourth, simulation omits real radio congestion, correlated loss, and timing
jitter. A deployment should preserve application bits for semantic comparison
and add wire bits, queue delay, and burst loss.

Further experiments should target policies cheaper than the confirmed
no-release points in baseline and scaled operation, and the rich-status point
in lossy operation. A failure to confirm sufficiency does not establish
insufficiency; fresh paired tests are needed to resolve these cheaper
candidates. Each adaptive round needs a new seed block and a predeclared error
budget, or an anytime-valid procedure.
Sensitivity analyses can repeat the final rule at $\delta=.02$ and $.10$, while
retaining $.05$ as the primary margin.

\section{Reproducibility}

The supplied release contains the experiment source, exact configuration,
per-episode compressed CSV files, stage selections, scenario seeds, analysis
scripts, tests and figure-generation code. It is available at
\url{https://github.com/sherryjiang-hub/Multi-robot-collaboration-}.

\section{Conclusion}

We present a method for assessing sufficient communication in independent BT
teams. It varies an explicit protocol family and message-admission gate,
measures actual recipient-weighted traffic, and separates exploratory selection
from fresh paired confirmation. Simultaneous confidence bounds support the
listed sufficient policies, while the claim rules prevent a selected frontier
crossing from being mistaken for a positive minimum. The zero-communication
result follows from unique task capabilities and the simulator's independent
task structure. For contested conditions, the reported operating points quantify
what the specified controller and protocols can achieve; each new task condition
requires its own evidence.

\paragraph{AI-use disclosure.}
The authors used OpenAI Codex for language
editing, code assistance, and workflow automation. The authors reviewed all
generated material and take responsibility for the ideas, claims, experiments,
results, and text.

\end{document}